\documentclass[letterpaper]{article} 
\usepackage{aaai2026}  
\usepackage{times}  
\usepackage{helvet}  
\usepackage{courier}  
\usepackage[hyphens]{url}  
\usepackage{graphicx} 
\usepackage{natbib}  
\usepackage{caption} 
\usepackage{algorithm}
\usepackage{algorithmic}
\usepackage{amsmath}
\usepackage{amssymb}
\usepackage{booktabs} 
\usepackage{multirow}
\usepackage{graphicx}
\usepackage{subcaption}
\usepackage{float}
\usepackage{xcolor}
\usepackage{newfloat}
\usepackage{listings}
\DeclareCaptionStyle{ruled}{labelfont=normalfont,labelsep=colon,strut=off} 
\floatstyle{ruled}
\newfloat{listing}{tb}{lst}{}
\floatname{listing}{Listing}
\nocopyright 

\title{Beyond Illumination: Fine-Grained Detail Preservation in Extreme Dark Image Restoration}
\author{
    Tongshun Zhang\textsuperscript{\rm 1,\rm 2},
    Pingping Liu\textsuperscript{\rm 1,\rm 2}\thanks{Corresponding author},
    Zixuan Zhong\textsuperscript{\rm 3},
    Zijian Zhang\textsuperscript{\rm 1,\rm 2},
    Qiuzhan Zhou\textsuperscript{\rm 4}
}
\affiliations{
    \textsuperscript{\rm 1} College of Computer Science and Technology, Jilin University\\
    \textsuperscript{\rm 2} Key Laboratory of Symbolic Computation and Knowledge Engineering of Ministry of Education, Jilin University\\
    \textsuperscript{\rm 3} College of Software, Jilin University\\
    \textsuperscript{\rm 4} College of Communication Engineering, Jilin University\\

    \{tszhang23, zhongzx24\}@mails.jlu.edu.cn,\{liupp, zhangzijian, zhouqz\}@jlu.edu.cn
}

\usepackage{bibentry}

\begin{document}

\maketitle

\begin{abstract}
Recovering fine-grained details in extremely dark images remains challenging due to severe structural information loss and noise corruption. Existing enhancement methods often fail to preserve intricate details and sharp edges, limiting their effectiveness in downstream applications like text and edge detection. To address these deficiencies, we propose an efficient dual-stage approach centered on detail recovery for dark images. In the first stage, we introduce a Residual Fourier-Guided Module (RFGM) that effectively restores global illumination in the frequency domain. RFGM captures inter-stage and inter-channel dependencies through residual connections, providing robust priors for high-fidelity frequency processing while mitigating error accumulation risks from unreliable priors. The second stage employs complementary Mamba modules specifically designed for textural structure refinement: (1) Patch Mamba operates on channel-concatenated non-downsampled patches, meticulously modeling pixel-level correlations to enhance fine-grained details without resolution loss. (2) Grad Mamba explicitly focuses on high-gradient regions, alleviating state decay in state space models and prioritizing reconstruction of sharp edges and boundaries. Extensive experiments on multiple benchmark datasets and downstream applications demonstrate that our method significantly improves detail recovery performance while maintaining efficiency. Crucially, the proposed modules are lightweight and can be seamlessly integrated into existing Fourier-based frameworks with minimal computational overhead. Code is available at https://github.com/bywlzts/RFGM.
\end{abstract}

\begin{figure}[ht]
    
    \centering  
    \begin{subfigure}[t]{\linewidth}  
        \centering  
        \includegraphics[width=0.90\linewidth]{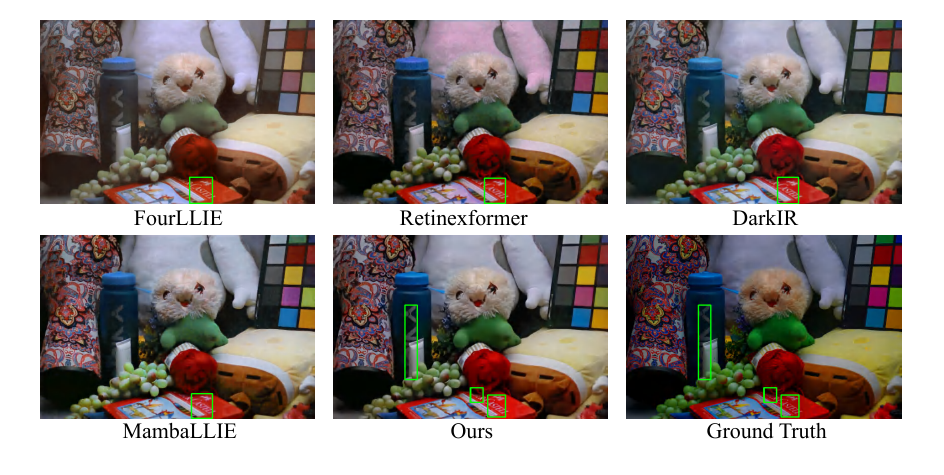}
        \vspace{-0.2cm}
        \caption{}
        \vspace{-0.1cm}
    \end{subfigure}  
    \begin{subfigure}[t]{\linewidth}  
        \centering  
        \includegraphics[width=0.90\linewidth]{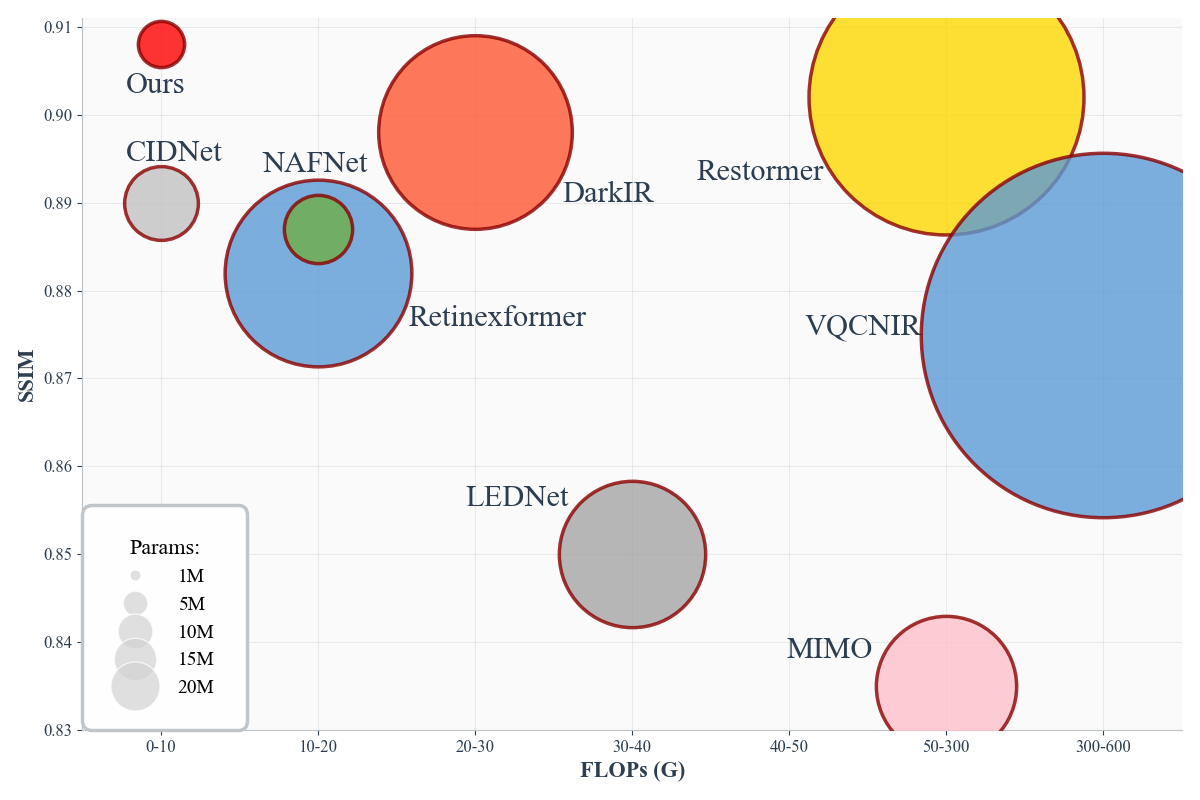}
        \vspace{-0.3cm}
        \caption{}
        \vspace{-0.3cm}
    \end{subfigure}  
    \caption{(a) Text detection comparisons on LOL-v1 dataset. (b) SSIM vs. Computation overhead on LOL-Blur dataset.}
    \vspace{-0.3cm}
    \label{fig:page1}  
\end{figure}

\section{Introduction}
Images captured in extremely dark conditions often suffer from poor visibility, leading to significant loss of structural and detailed information, which constrains the performance of fine-grained downstream applications. Traditional restoration techniques, such as histogram equalization~\cite{pizer1990contrast}, Retinex theory~\cite{guo2016lime}, and gamma correction~\cite{rahman2016adaptive}, have been explored but struggle in extreme darkness and exhibit limited generalization, leading to their decline. Recently, learning-based methods~\cite{zou2024vqcnir, weng2025mamballie, feijoo2025darkir} have improved reconstruction quality and scene generalization by learning mappings between low-light and normal-light images. However, these approaches primarily focus on global brightness mapping, failing to preserve fine details effectively. They are also hampered by limited noise control and complex architectures with large parameter counts, making it difficult to balance performance and efficiency.

To address the computational complexity and parameter efficiency challenges inherent in spatial-domain methods, frequency-domain approaches~\cite{four2,four1,UHDFourICLR2023,zhang2024dmfourllie,feijoo2025darkir} have emerged as a promising alternative for dark image restoration. While Fourier-domain methods can achieve effective global information modeling while maintaining compact parameter specifications, these methods commonly rely on sequential simple convolutions or introduce unreliable priors, which lead to redundancy or loss of frequency-domain information. Furthermore, due to the global modeling nature of the Fourier domain, these methods employ additional encoder-decoder structures to enhance spatial detail representation. However, encoder-decoder architectures struggle to capture fine-grained structure and details, and the downsampling process inevitably leads to the loss of critical image detail information.

Concurrently, Mamba~\cite{gu2023mamba} based dark image restoration~\cite{zou2024wave, bai2024retinexmamba, weng2025mamballie} methods have attracted significant attention due to their linear complexity, with Mamba demonstrating tremendous potential in balancing global receptive fields and computational efficiency. However, these methods unfold 2D images using fixed scanning rules to generate 1D token sequences, introducing redundancy through multiple redundant scans. Moreover, tokens with strong associative properties may be spatially distant in the sequence~\cite{guo2025mambair}, thereby weakening inter-token modeling capabilities and limiting long-range dependency modeling.

Motivated by these critical limitations, we propose an efficient dual-stage approach specifically designed for dark image detail recovery. Our method strategically addresses the aforementioned challenges through the synergistic combination of frequency-domain global modeling and spatial-domain detail refinement.
\textbf{In the first stage,} we introduce a Residual Fourier-Guided Module (RFGM). In the Fourier domain, the amplitude component represents the brightness of an image, while the phase component encodes its structural details. Amplitude recovery requires precise amplitude mapping, whereas phase components necessitate robust adaptive adjustment. Therefore, we leverage inter-stage and inter-channel correlations in a residual manner to provide robust prior guidance. We identify optimally matched amplitude components as residual Fourier channels, serving as prior guidance for amplitude mapping in subsequent stages, while phase components provide additional structural prior compensation for later stages in a residual fashion. This achieves efficient and robust recovery of global frequency-domain information.
\textbf{In the second stage,} we advance beyond illumination to prioritize fine-grained detail preservation. To achieve this, we propose complementary dual-branch Mamba modules that work synergistically: Patch Mamba specializes in pixel-level fine detail enhancement, while Grad Mamba targets the reconstruction of structural textures. Patch Mamba functions on channel-concatenated, non-downsampled patches, avoiding the pitfalls of encoding-decoding sampling losses and meticulously modeling pixel-level correlations to enhance fine details without sacrificing resolution or increasing computational load. In contrast, Grad Mamba concentrates on high-gradient regions, utilizing gradient score prediction to enhance interactions among tokens that are closely associated with gradients. Inspired by MambaIRv2~\cite{guo2025mambairv2}, we further integrate gradient prediction scores with state space models to alleviate state decay, ensuring a concentrated effort on reconstructing sharp edges and boundaries.

In Fig. \ref{fig:page1}, we validate the effectiveness of our method for text detection and denoising, showcasing its capacity to restore fine details in extremely dark images. 
In summary, our main contributions are as follows:  
\vspace{-0.1cm}
\begin{itemize}
\item We propose an efficient dark image restoration framework focused on fine-grained detail preservation.
\item We present a Residual Fourier-Guided Module (RFGM), which utilizes inter-stage and inter-channel correlations to enhance prior guidance and mitigate issues of redundancy and error propagation.
\item We overcome Mamba's inherent limitations by designing dual-branch modules: Patch Mamba for fine detail enhancement without resolution loss, and Grad Mamba for gradient-driven structural boundary reconstruction.
\item Extensive experiments demonstrate significant improvements in restoration quality for dark images and downstream tasks (text detection, edge detection) while maintaining minimal computational overhead and compatibility with existing Fourier-based methods.
\end{itemize}

\begin{figure*}[t]
    \centering
    \includegraphics[width=1\linewidth]{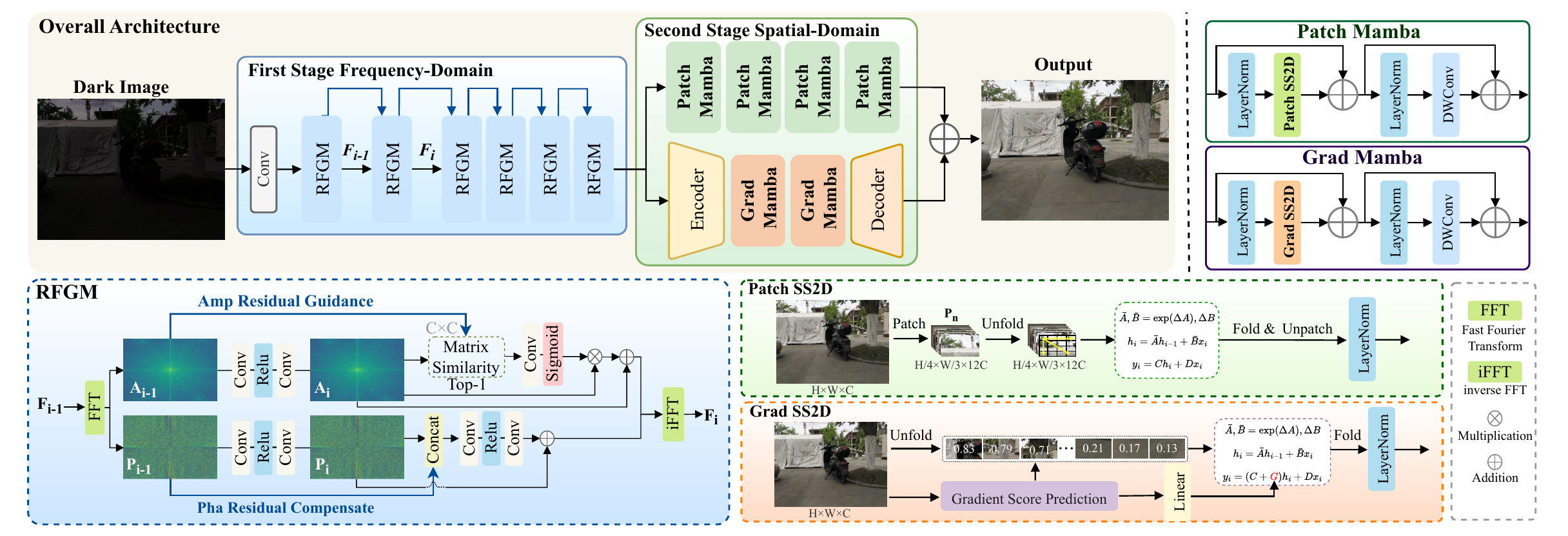}
    \vspace{-0.3cm}
    \caption{Overall architecture of our proposed dual-stage framework. The First Stage Frequency Domain comprises six RFGMs, while the Second Stage Spatial Domain consists of four Patch Mambas and encoder-decoder with two Grad Mambas.}
    \vspace{-0.2cm}
    \label{fig:network}
\end{figure*}

\section{Related Work}

\textbf{Frequency-Based Dark Image Restoration Methods.}
Frequency-domain approaches~\cite{four1, four2, feijoo2025darkir} have proven effective by distinguishing high-frequency from low-frequency information, enhancing brightness while minimizing noise. FourLLIE~\cite{four1} leverages Fourier transforms for efficient global feature extraction, replacing Transformer modules in SNR-Aware~\cite{lowlight8} and significantly reducing parameter counts. UHDFour~\cite{UHDFourICLR2023} enhances ultra-high-definition images by utilizing consistent amplitude patterns across resolutions but suffers from information loss. DMFourLLIE~\cite{zhang2024dmfourllie} enhances frequency-domain information by introducing infrared priors, but fails to consider the generalization limitations of pretrained models. Wavelet-based methods, such as Wave-Mamba~\cite{zou2024wave}, apply wavelet transforms in ultra-high-definition enhancement but face challenges due to complexity in low-frequency processing architectures~\cite{jiang2023low}. CWNet~\cite{zhang2025cwnet} combines causal and wavelet methods for brightness restoration but neglects detailed modeling of image nuances. 

\noindent \textbf{Mamba-Based Dark Image Restoration Methods.}
Mamba~\cite{gu2023mamba} introduced input-dependent state space models (SSMs) with selective mechanisms, applied successfully across various tasks like super-resolution~\cite{guo2025mambairv2}, classification~\cite{xiao2024spatial}, and restoration~\cite{li2025mair}. In low-light enhancement, RetinexMamba~\cite{bai2024retinexmamba} utilized Mamba with Retinex theory for improved efficiency. Wave-Mamba~\cite{zou2024wave} integrated Mamba with wavelet transforms, while MambaLLIE~\cite{weng2025mamballie} introduced implicit Retinex-aware mechanisms in a state space model. However, these methods do not resolve the limitations of state space models in 2D applications, which hampers effective token modeling. Our work not only addresses the decay of state space models but also pioneers pixel-level fine-grained modeling with Mamba.

\begin{figure}[t]
    \centering
    \includegraphics[width=0.8\linewidth]{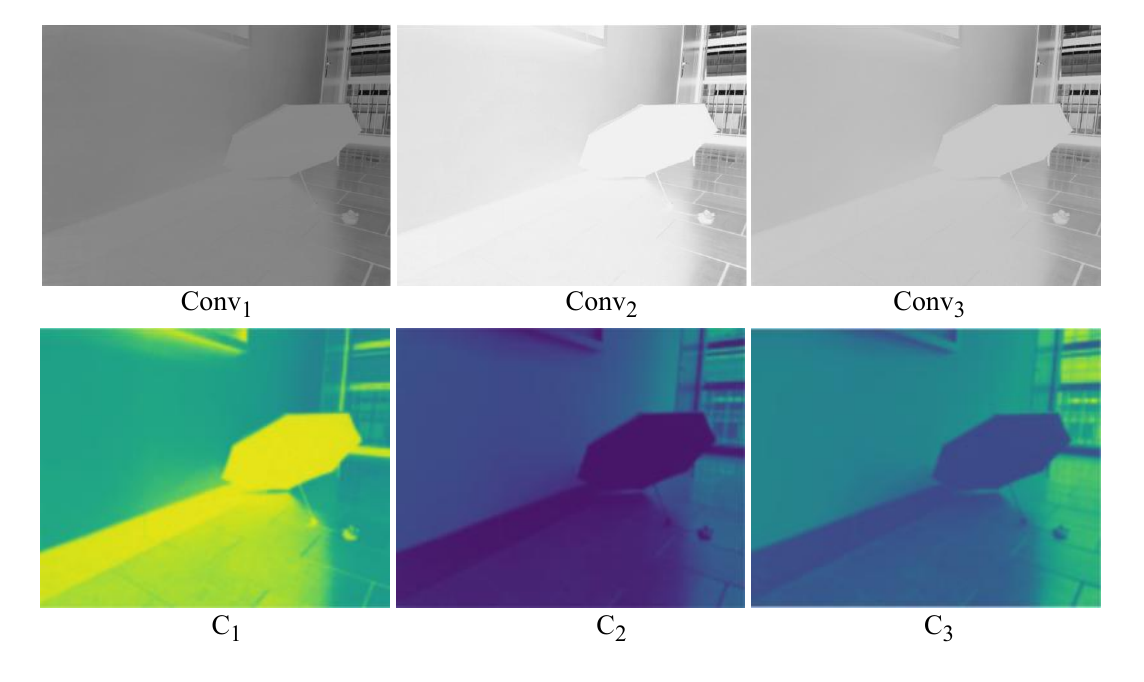}
    \vspace{-0.2cm}
    \caption{ Visualization of DMFourLLIE across different stages and channels. $\text{Conv}_i$ denotes stage-specific convolutional features, while $\text{C}_i$ represents distinct feature channels.}
    \vspace{-0.2cm}
    \label{fig:motivation1}
\end{figure}

\section{Method}

\subsection{Overview}  
The overall architecture of our proposed dual-stage approach is illustrated in Fig.~\ref{fig:network}. Given a dark image $I \in \mathbb{R}^{H \times W \times 3}$, we first apply a $3 \times 3$ convolutional layer to extract shallow feature embeddings of size $\mathbb{R}^{H \times W \times C}$, where $H$, $W$, and $C$ denote height, width, and channel dimensions, respectively. \textbf{First Stage - Frequency-Domain Global Modeling:} Global information is processed through six identical RFGMs, which leverage inter-stage and inter-channel correlations in a residual manner to provide robust prior guidance. This stage focuses on global illumination recovery and overall structural restoration. \textbf{Second Stage - Spatial-Domain Detail Refinement:} Decoupled from illumination adjustment, the second stage employs complementary dual-branch Mamba modules that work synergistically to preserve fine-grained textural and structural details. One branch consists of four Patch Mamba modules operating on channel-concatenated non-downsampled image patches, ensuring computational efficiency while precisely modeling pixel-level correlations. The other branch comprises an encoder-decoder and two Grad Mamba modules, specifically designed for structural texture reconstruction with emphasis on sharp edge and boundary recovery.

\subsection{First Stage - Frequency-Domain Global Modeling}

\noindent \textbf{Motivation Analysis.} As visualized in Fig.~\ref{fig:motivation1}, current methods~\cite{four2, zhang2024dmfourllie} typically handle amplitude and phase components sequentially through a series of convolutional blocks (Conv$_1 \rightarrow$ Conv$_2 \rightarrow$ Conv$_3$). However, our feature analysis uncovers inconsistencies in feature representations across these stages. Notably, the features from Conv$_2$ exhibit significantly brighter activations compared to those from Conv$_3$. This inconsistency signals potential information degradation inherent in a strictly sequential processing paradigm, where vital cues from earlier stages (Conv$_2$) are insufficiently preserved or leveraged in subsequent stages (Conv$_3$), leading to less effective illumination adjustments.
Moreover, a channel-wise examination of features (C$_1$, C$_2$, C$_3$) within a single stage reveals that essential structural contours manifest with varying prominence across different channels (C$_1$ vs. C$_2$ vs. C$_3$). This disparity emphasizes the limitation of treating channels in isolation, suggesting a failure to synthesize the complementary structural information that is distributed across the channels.
Collectively, Fig.~\ref{fig:motivation1} elucidates two critical shortcomings: (1) the risk of progressive information loss due to sequential stage processing, and (2) the oversight of comprehensive structural synthesis resulting from isolated channel processing. These deficiencies underscore the urgent need for strategies that explicitly foster inter-stage information preservation and inter-channel information fusion.

Motivated by the observation, and to eliminate error accumulation from pre-trained model~\cite{wu2023learning,zhang2024dmfourllie} and manual priors~\cite{bai2024retinexmamba} while reducing computational overhead from additional modules~\cite{xu2023low}, we propose the Residual Fourier-Guided Module (RFGM). This module captures the most valuable amplitude channel priors from the previous stage, avoiding redundant processing while providing precise guidance for amplitude component mapping in subsequent stages. Meanwhile, phase components provide robust structural compensation through residual connections, enabling adaptive structural information reconstruction.

Specifically, as shown in the bottom-left of Fig.~\ref{fig:network}, features $F_{i\text{-}1} \in \mathbb{R}^{H \times W \times C}$ from the previous stage are first transformed to the Fourier domain using Fast Fourier Transform (FFT), yielding amplitude $A_{i\text{-}1}$ and phase $P_{i\text{-}1}$ components. These components are processed separately: $A_{i\text{-}1}$ and $P_{i\text{-}1}$ undergo convolution followed by ReLU activation to obtain $A_{i}$ and $P_{i}$, respectively.
For amplitude components, both $A_{i-1}$ and $A_{i}$ are flattened to $\mathbb{R}^{HW \times C}$ and subjected to matrix similarity computation $MS(\cdot, \cdot)$, yielding a similarity matrix $M \in \mathbb{R}^{C \times C}$. From $M$, we select a Top-1 vector $V \in \mathbb{R}^{C \times 1}$ as an index corresponding to the most similar channel between $A_{i-1}$ and $A_{i}$. The objective is to select the most reliable brightness prior from channels with varying brightness distributions in $A_{i-1}$. The selected prior information is then expanded through $1 \times 1$ convolution and processed with a Sigmoid activation function to generate prior guidance $P_{a}$. Subsequently, amplitude $A_{i}$ is multiplied by $P_{a}$ and combined with residual connections to complete the prior guidance fusion from the previous stage. This process can be formulated as:
\vspace{-0.05cm}
\begin{equation}  
\begin{aligned}  
\begin{gathered}  
M = MS(A_{i-1}, A_{i}), \quad
V = \text{Top-1}(M),\\
P_{a} = \text{Sigmoid}(\text{Conv}(A_{i-1}|\text{Index}(V))),\\
\tilde{A_{i}} = A_{i} \times P_{a} + A_{i}.  
\end{gathered}  
\end{aligned}  
\end{equation}

For phase components, $P_{i-1}$ and $P_{i}$ are concatenated along the channel dimension, where phase information from the previous stage serves as structural compensation and is adaptively fused through convolutional layers. Subsequently, through ReLU activation and convolution, the phase is restored to its original scale $\tilde{P_{i}} \in \mathbb{R}^{H \times W \times C}$:
\vspace{-0.05cm}
\begin{equation}  
\tilde{P_{i}} = \text{Conv}(\text{Concat}(P_{i-1}, P_{i})) + P_{i}.
\end{equation} 

Finally, the processed amplitude $\tilde{A_{i}}$ and phase $\tilde{P_{i}}$ are combined through inverse Fast Fourier Transform (iFFT) to generate the output features $F_{i}$ for the next stage.

\subsection{Second Stage - Spatial-Domain Detail Refinement}
\noindent \textbf{Motivation Analysis.} While frequency-domain illumination recovery effectively tackles low-frequency brightness challenges in dark images, it is essential to focus on fine-grained structural detail and sharp edge preservation, surpassing mere illumination adjustment. Therefore, we introduce complementary dual-branch Mamba aimed at fine-grained detail recovery.
Mamba-based restoration methods employ discrete state space equations to model interactions between tokens:
\vspace{-0.05cm}
\begin{equation}
h_i = \overline{\mathbf{A}} h_{i-1} + \overline{\mathbf{B}} x_i, \quad
y_i = \mathbf{C} h_i + \mathbf{D} x_i,
\end{equation}
where the $i$-th token depends entirely on the preceding $i\text{-}1$ tokens, creating direct causal relationships between neighboring pixels. However, this scanning mechanism disrupts correlations among distant features, rendering it less effective for vision tasks. Furthermore, the causal modeling may induce long-range decay effects, while multiple fixed-direction scanning strategies add unnecessary complexity and information redundancy~\cite{guo2025mambairv2}.

Based on these observations, rather than being constrained by Mamba's limitations, we reverse the approach by fully leveraging Mamba's unique characteristics and transforming them into advantages for 2D image modeling.

\begin{table*}[t]  
\centering  
\resizebox{\textwidth}{!}{  
\begin{tabular}{lccc|ccc|ccc}  
\toprule  
\multirow{2}{*}{Methods} & \multicolumn{3}{c|}{LOL-v1} & \multicolumn{3}{c|}{LOL-v2-Real} & \multicolumn{3}{c}{LOL-v2-Syn} \\   
\cmidrule(lr){2-4} \cmidrule(lr){5-7} \cmidrule(lr){8-10}  
& PSNR ↑ & SSIM ↑ & LPIPS ↓ & PSNR ↑ & SSIM ↑ & LPIPS ↓ & PSNR ↑ & SSIM ↑ & LPIPS ↓ \\   
\midrule   
Kind~\cite{kind} & 20.87 & 0.7995 & 0.2071 & 20.01 & 0.8412 & 0.0813 & 22.62 & 0.9041 & 0.0515 \\   
MIRNet~\cite{lowlight9} & 24.14 & 0.8305 & 0.2502 & 22.11 & 0.7942 & 0.1448 & 22.52 & 0.8997 & 0.0568 \\   
Kind++~\cite{kind++} & 18.97 & 0.8042 & 0.1756 & 20.59 & 0.8294 & 0.0875 & 21.17 & 0.8814 & 0.0678 \\    
SNR-Aware~\cite{lowlight8} & 23.93 & 0.8460 & 0.0813 & 21.48 & 0.8478 & 0.0740 & 24.13 & 0.9269 & 0.0318 \\   
FourLLIE~\cite{four1} & 20.99 & 0.8071 & 0.0952 & 22.34 & 0.8403 & 0.0573 & 24.65 & 0.9192 & 0.0389 \\   
UHDFour~\cite{UHDFourICLR2023} & 22.89 & 0.8147 & 0.0934 & 19.42 & 0.7896 & 0.1151 & 23.64 & 0.8998 & 0.0341 \\   
Retinexformer~\cite{retinexformer} & 22.71 & 0.8177 & 0.0922 & 22.79 & 0.8397 & 0.0724 & 25.67 & 0.9295 & 0.0273 \\    
DMFourLLIE~\cite{zhang2024dmfourllie} & 22.98 & 0.8273 & 0.0792 & 22.71 & 0.8583 & 0.0539 & 25.74 & 0.9308 & 0.0251 \\  
UHDFormer~\cite{wang2024uhdformer} & 22.88 & 0.8370 & 0.1390 & 19.71 & 0.8320 & 0.0758 & 24.48 & 0.9277 & 0.0306 \\  
Wave-Mamba~\cite{zou2024wave} & 22.76 & 0.8419 & 0.0791 & 20.35 & 0.8379 & 0.1908 & 24.69 & 0.9271 & 0.0584 \\    
RetinexMamba~\cite{bai2024retinexmamba} & 23.15 & 0.8210 & 0.0876 & 21.73 & 0.8290 & 0.1164 & 25.89 & 0.9346 & 0.0389 \\  
MambaLLIE~\cite{weng2025mamballie}& 22.80 & 0.8315 &0.0907 & 21.85 & 0.8276 & 0.1673 & \underline{25.87} & \underline{0.9400} & 0.0467 \\ 
CWNet~\cite{zhang2025cwnet} & \underline{23.60} & 0.8496 & \underline{0.0648} & 23.31 & \underline{0.8641} & \underline{0.0532} & 25.74 & 0.9365 & \underline{0.0241}\\  
CIDNet~\cite{yan2025hvi} & 23.81 & \textbf{0.8574} & 0.0856 & \textbf{23.43} & 0.8622 & 0.1691 & 25.70 & 0.9419 & 0.0437 \\  
URetienxNet++~\cite{yan2025hvi} & 23.83 &0.8390 &0.2310 & 21.97 & 0.8360 & 0.2030 & 24.60 & 0.9270 & 0.1020 \\  
\midrule  
Ours & \textbf{24.11} & \underline{0.8517} & \textbf{0.0557} & \underline{23.38} & \textbf{0.8662} & \textbf{0.0527} & \textbf{25.97} & \textbf{0.9408} & \textbf{0.0195} \\  
\bottomrule  
\end{tabular}  
}  
\vspace{-0.1cm}
\caption{Quantitative comparison on LOL-v1, LOL-v2-Real, and LOL-v2-Syn~\cite{lol} datasets without using ground truth mean. The best and second-best results are highlighted in \textbf{bold} and \underline{underlined}, respectively. }  
\vspace{-0.4cm}
\label{tab:com}  
\end{table*}

\begin{table}[t]  
\centering  
\renewcommand{\arraystretch}{1.1}
\setlength{\tabcolsep}{3pt}
\resizebox{0.5\textwidth}{!}{ 
\begin{tabular}{l|cc|cc|cc}  
\toprule  
\multirow{2}{*}{Methods} & \multicolumn{2}{c|}{LSRW-Huawei} & \multicolumn{2}{c|}{LSRW-Nikon} & \multicolumn{2}{c}{Efficiency} \\   
                 & PSNR ↑ & SSIM ↑ & PSNR ↑ & SSIM ↑ & \#Param (M) & \#FLOPs (G) \\ \midrule  
Kind & 16.58 & 0.5690 &11.52 &0.3827 & 8.02 & 34.99 \\
MIRNet & 19.98 & 0.6085 & 17.10 & 0.5022 & 31.79 & 785.1 \\
SNR-Aware & 20.67 & 0.5911 & 17.54 & 0.4822 & 39.12 & 26.35 \\
UHDFour & 19.39 & 0.6006 & \underline{17.94} & 0.5195 & 17.54 & \underline{4.78} \\
Retinexformer & 21.23 & 0.6309 & 17.64 & 0.5082 & 1.61 & 15.57 \\ 
Wave-Mamba & 21.19 & 0.6391 & 17.34 & 0.5192 & 1.26 & 7.22 \\ 
DMFourLLIE & 21.47 & 0.6331 & 17.04 & \underline{0.5274} & \underline{0.75} & 5.81 \\ 
Retinexmamba & 20.88 & 0.6298 & 17.59 & 0.5133 & 3.59 & 34.76 \\ 
CWNet & \underline{21.50} & \underline{0.6397} & 17.38 & 0.5119 & 1.23 & 11.3 \\
CIDNet & 20.30 & 0.6054 & 17.16 & 0.4975 & 1.88 & 7.57 \\ 
MambaLLIE & 20.98 & 0.6388 & 17.25 & 0.5084 & 2.28 & 20.85 \\ 
 \midrule
Ours & \textbf{21.59} & \textbf{0.6441} & \textbf{17.98} & \textbf{0.5357} & \textbf{0.37} & \textbf{4.34} \\ 
\bottomrule  
\end{tabular}  }
\vspace{-0.1cm}
\caption{Quantitative comparison on LSRW-Huawei and LSRW-Nikon datasets.}  
\vspace{-0.1cm}
\label{tab:com-huawei}  
\end{table}

\noindent \textbf{Patch Mamba.}
Image details are manifested in pixel-level variations and correlations. However, direct pixel-level operations incur substantial memory overhead. While encoder-decoder architectures can enhance spatial performance, sampling layers inevitably lose crucial pixel-level content, which is devastating for fine-grained detail preservation. Since we focus primarily on relationships between adjacent pixels, capturing global contextual dependencies is unnecessary. Mamba's robust scanning mechanism precisely satisfies pixel adjacency requirements while weakening connections between distant pixels. However, processing without downsampling creates computational burden and poor parallelization. Therefore, we propose Patch Mamba, which operates on channel-concatenated non-downsampled patches. As illustrated in the top-right of Fig.~\ref{fig:network}, Patch Mamba consists of two residual blocks: the first comprising LayerNorm and Patch SS2D, and the second comprising LayerNorm and depthwise separable convolution.

Patch SS2D performs patch-wise selective scanning. Given an input feature map $F \in \mathbb{R}^{H \times W \times C}$, it is uniformly divided into $(n,m)$ non-overlapping patches $P_{i,j} \in \mathbb{R}^{h \times w \times C}$, where $h = \frac{H}{\sqrt{n}}$ and $w = \frac{W}{\sqrt{m}}$. These patches are then stacked along the channel dimension to form $\mathbb{R}^{h \times w \times n \cdot m \cdot C}$, significantly reducing scanning dimensions and alleviating computational burden while improving parallel efficiency. Note that the Patch SS2D scanning strategy encompasses horizontal, vertical, diagonal, and their respective reverse directions to comprehensively enhance inter-pixel correlations. After scanning all patches, results are recombined into $\mathbb{R}^{H \times W \times C}$ for subsequent processing.

\noindent \textbf{Grad Mamba.}
Previous methods often fail to preserve edge sharpness and textural integrity, especially in areas with complex gradient distributions. To address this, we propose Grad Mamba, a gradient-guided state space model that explicitly targets high-gradient regions and prioritizes the reconstruction of sharp edges and boundaries.

As shown in Fig.~\ref{fig:network}, Grad Mamba operates within an encoder-decoder framework, complementing Patch Mamba. Its overall structure is similar to Patch Mamba. Specifically, given input features $F \in \mathbb{R}^{H \times W \times C}$, we first perform gradient score prediction using three complementary operators: Sobel operators in x and y directions to capture directional gradients, and a Laplacian operator to detect edge transitions. The extracted gradient magnitudes are then converted into priority scores through adaptive normalization and learnable scaling parameters:
\vspace{-0.05cm}
\begin{equation}
P_{grad} = \sigma(\cdot \frac{G_{mag} - G_{min}}{G_{max} - G_{min} + \epsilon} + \beta),
\end{equation}
where $P_{grad} \in \mathbb{R}^{H \times W}$ represents gradient priority scores, $\beta$ is learnable offset parameter, $\sigma$ denotes the sigmoid function, and $G_{min}$, $G_{max}$ are the minimum and maximum gradient magnitudes for each sample.

Given gradient priority scores $P_{grad}$, we sort tokens according to their gradient importance, prioritizing high-gradient tokens for processing. This enables preferential interactions between high-gradient associated regions while avoiding long-range decay limitations. Additionally, inspired by MambaIRv2~\cite{guo2025mambairv2}, since state $C$ in state space models resembles the query $Q$ in attention mechanisms, it acquires attention-like capabilities to query high-gradient regions throughout the image, making the network more focused on edge information and structural detail reconstruction.
Therefore, the enhanced state space formulation is expressed as:
\vspace{-0.03cm}
\begin{equation}
h_i = \overline{\mathbf{A}} h_{i-1} + \overline{\mathbf{B}} x_i, \quad
y_i = (\mathbf{C} + \mathbf{G}) h_i + \mathbf{D} x_i,
\end{equation}
where $\mathbf{G} \in \mathbb{R}^{N \times D}$ represents the gradient guidance matrix derived from $P_{grad}$, and is computed as:
\vspace{-0.03cm}
\begin{equation}
\mathbf{G} = \text{Linear}(P_{grad}) \cdot \mathbf{W_G},
\end{equation}
where $\mathbf{W_G}$ is a learnable projection matrix that transforms gradient scores into the same dimensional space as the output projection matrix $\mathbf{C}$. In high-gradient regions, the enhanced projection $(\mathbf{C} + \mathbf{G})$ amplifies the hidden state contributions, effectively strengthening the model's focus on edge and structural information. Through this mechanism, Grad Mamba achieves adaptive enhancement of structural details while maintaining the computational efficiency and sequential modeling capabilities inherent to state space models.

\begin{table}[t]  
\centering  
\renewcommand{\arraystretch}{1.1}
\setlength{\tabcolsep}{3pt}
\resizebox{0.5\textwidth}{!}{ 
\begin{tabular}{l|ccc|cc}  
\toprule  
\multirow{2}{*}{Methods} & \multicolumn{3}{c|}{LOL-Blur} & \multicolumn{2}{c}{Efficiency} \\   
                 & PSNR ↑ & SSIM ↑ & LPIPS ↓ & \#Param (M) & \#FLOPs (G) \\ \midrule  
DRBN & 21.78 & 0.768 & 0.325 & \underline{0.6} & 48.61 \\
DeblurGAN-v2 & 20.30 & 0.745 & 0.356 & 60.9 & - \\
MIMO & 22.41 & 0.835 & 0.262 & 6.8 & 67.25 \\
NAFNet & 25.36 & 0.882 & 0.158 & 12.05 & 12.3 \\
LEDNet* & 25.74 & 0.850 & 0.224 & 7.4 & 38.65 \\
Retinexformer & 26.02 & 0.887 & 0.181 & 1.61 & 15.57 \\ 
Restormer & 26.72 & 0.902 & 0.133 & 26.13 & 144.25 \\ 
VQCNIR*   & \textbf{27.79} &  0.875 & \textbf{0.096} & 45.9 & 325.27 \\
DarkIR* & 27.30 & 0.898 & 0.137 &12.96 & 27.19 \\  
CIDNet & 26.57 &0.890 & 0.120 & 1.88 & \underline{7.57} \\ 
\midrule
Ours & \underline{27.49} & \textbf{0.908} & \underline{0.105} & \textbf{0.37} & \textbf{4.34} \\
\bottomrule  
\end{tabular}  }
\vspace{-0.1cm}
\caption{Quantitative comparison on LOL-Blur dataset. The methods marked with * are specifically designed for joint denoising and restoration.}  
\vspace{-0.1cm}
\label{tab:com-lolblur}  
\end{table}

\begin{figure*}[t]
    \centering
    \includegraphics[width=1\linewidth]{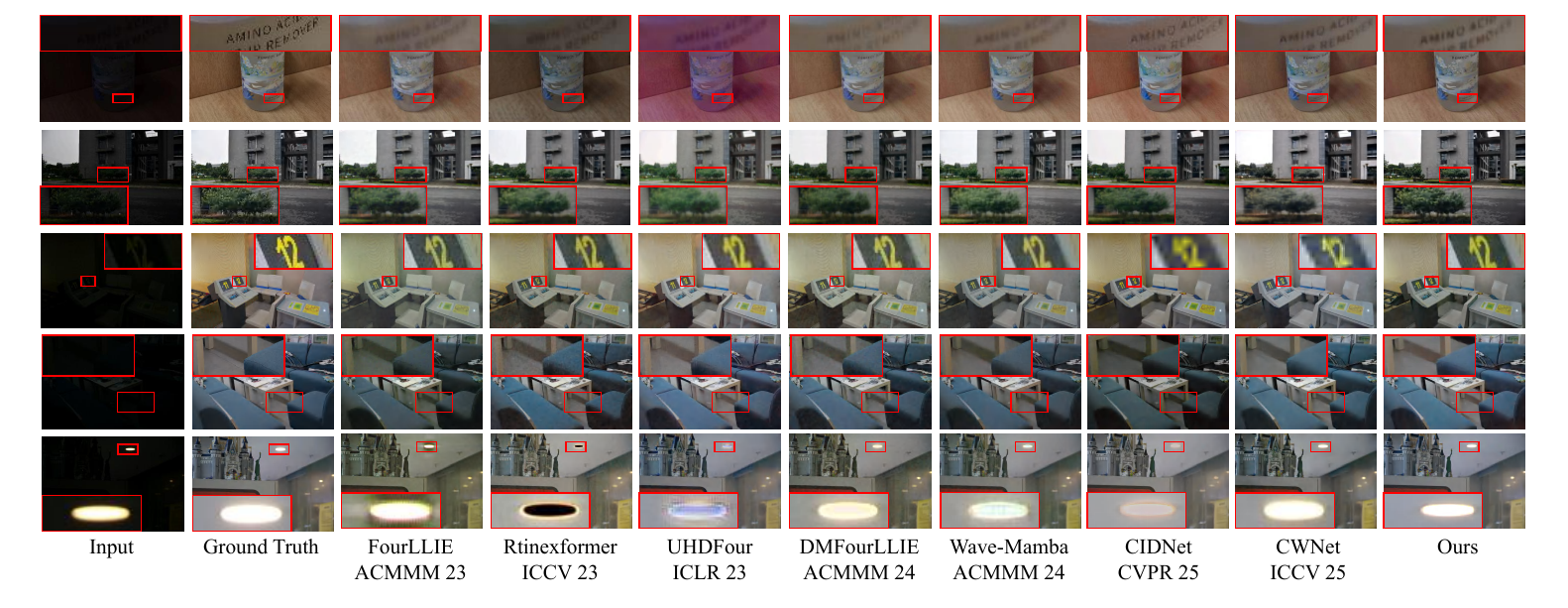}
    \vspace{-0.6cm}
    \caption{Visual comparisons with state-of-the-art methods. From top to bottom: LSRW-Huawei (row 1), LSRW-Nikon (row 2), LOL-v1 (row 3), and LOL-v2-Real (rows 4-5) datasets.}
    \vspace{-0.4cm}
    \label{fig:visual_comparison}
\end{figure*}

\begin{table}[t]  
\centering  
\renewcommand{\arraystretch}{1.2} 
\setlength{\tabcolsep}{4pt} 
\resizebox{0.80\columnwidth}{!}{ 
\begin{tabular}{l|ccc}  
\toprule  
\textbf{Methods} & \textbf{PSNR ↑} & \textbf{SSIM ↑} & \textbf{LPIPS ↓} \\ \midrule 
w/o First Stage        & 20.73           & 0.6331          & 0.1962          \\  
w/o Second Stage    & 20.95          & 0.6387          & 0.1827          \\  
w/o  RFGM   & 21.24           & 0.6374         & 0.1696          \\  
w/o Patch Mamba & 21.54           & 0.6427         & 0.1612          \\ 
w/o Grad Mamba    & 21.21           & 0.6392          & 0.1831          \\  
  \hline
Full Model      & \textbf{21.59}  & \textbf{0.6441} & \textbf{0.1607} \\  
\bottomrule  
\end{tabular}}  
\vspace{-0.2cm}
\caption{Ablation study on each component.}  
\vspace{-0.2cm}
\label{tab:ablationbranch}  
\end{table}

\section{Experiments}
\subsection{Datasets and Experimental Setting}
\noindent \textbf{Datasets.} We conduct comprehensive experiments on five benchmark datasets for extreme dark image restoration: LOL-v1~\cite{lol} (485 training/15 testing pairs), LOL-v2-Real (689 training/100 testing pairs), LOL-v2-Syn (900 training/100 testing pairs), LSRW-Huawei~\cite{lsrw} (3,150 training/20 testing pairs), LSRW-Nikon (2,450 training/30 testing pairs), and LOL-Blur~\cite{zhou2022lednet} (10,200 training/1,800 testing pairs). These datasets feature severely underexposed images captured in extremely low-light conditions, presenting significant challenges for meaningful visual content recovery.

\noindent \textbf{Implementation Details.}
Our method is implemented end-to-end on the PyTorch platform. Images are randomly cropped to $256 \times 256$ resolution and augmented with random horizontal/vertical flips and rotations. We use the ADAM optimizer with $\beta_1 = 0.9$ and $\beta_2 = 0.99$, initialized at a learning rate of $4.0 \times 10^{-4}$. Learning rate scheduling follows a MultiStepLR strategy with decay steps at $5 \times 10^{4}$ and $1 \times 10^{5}$ iterations, applying a decay factor of $0.5$. All experiments are conducted on dual NVIDIA RTX 4090 GPUs (24GB) and an Intel Core i9-14900K processor, with training performed using a batch size of 8 for $2 \times 10^{5}$ iterations. The experiments utilize ${\mathcal L}_{1}$ loss for training.

\begin{table}[t]  
\centering  
\renewcommand{\arraystretch}{1.2} 
\setlength{\tabcolsep}{4pt} 
\resizebox{\columnwidth}{!}{ 
\begin{tabular}{l|cccc}  
\toprule  
\textbf{Methods} & \textbf{PSNR ↑} & \textbf{SSIM ↑} & \textbf{LPIPS ↓}  & \textbf{Param (M)}\\ \midrule 
FourLLIE (Baseline)  & 21.11    & 0.6256   & 0.1825   & 0.12       \\  
FourLLIE + RFGM     & \underline{21.39}   & \underline{0.6425}   & \underline{0.1683}  & 0.126 (↓0.006)        \\
FourLLIE + RFGM + PatchSS2D     & \textbf{21.47}   & \textbf{0.6433}   & \textbf{0.1622}  & 0.14 (↓0.02)        \\  \midrule
DMFourLLIE (Baseline)  & 21.47    & 0.6331    & 0.1781   & 0.75       \\  
DMFourLLIE + RFGM      & \underline{21.50}     & \underline{0.6408}  & \underline{0.1593}   & 0.756 (↓0.06)         \\ 
DMFourLLIE + RFGM + PatchSS2D      & \textbf{21.52}     & \textbf{0.6439}  & \textbf{0.1492}   & 0.77 (↓0.02)         \\  

\bottomrule  
\end{tabular}}  
\vspace{-0.2cm}
\caption{Plug-and-play validation on Fourier-based methods.}  
\vspace{-0.2cm}
\label{tab:plug_test}  
\end{table}

\noindent \textbf{Comparative Methods and Evaluation Metrics.}
We compare our method against various SOTA approaches, including deep learning methods (Kind~\cite{kind}, Kind++~\cite{kind++}, MIRNet~\cite{lowlight9}, SNR-Aware~\cite{lowlight8}, CWNet~\cite{zhang2025cwnet}, CIDNet~\cite{yan2025hvi}, URetienxNet++~\cite{yan2025hvi}), Fourier-based methods (FourLLIE~\cite{four1}, UHDFour~\cite{UHDFourICLR2023}, DMFourLLIE~\cite{zhang2024dmfourllie}), Transformer-based methods (Retinexformer~\cite{retinexformer}, UHDFormer~\cite{wang2024uhdformer}), and Mamba-based methods (RetinexMamba~\cite{bai2024retinexmamba}, Wave-Mamba~\cite{zou2024wave}, MambaLLIE~\cite{weng2025mamballie}). All methods are evaluated using the same testing protocols for fairness, employing Peak Signal-to-Noise Ratio (PSNR), Structural Similarity Index (SSIM)~\cite{SSIM}, and Learned Perceptual Image Patch Similarity (LPIPS)~\cite{LP} as full-reference metrics.

\subsection{Quantitative and Qualitative Results}
\noindent \textbf{Comparison on LOL-v1, LOL-v2-Real, and LOL-v2-Syn Datasets.} Quantitative results are in Tab.~\ref{tab:com}. On the LOL-v1 dataset, our method achieves the highest PSNR and LPIPS while maintaining competitive SSIM. For the LOL-v2-Real dataset, we attain the second-best PSNR, the highest SSIM, and the best LPIPS score. On the LOL-v2-Syn dataset, our method outperforms all metrics.

\noindent \textbf{Comparison on LSRW-Huawei and LSRW-Nikon Datasets.} Results in Tab.~\ref{tab:com-huawei} show that we achieve the highest PSNR and SSIM on both datasets. \textbf{Notably}, our method is computationally efficient, requiring just \textbf{0.37M} parameters and \textbf{4.34G} FLOPs.

\noindent \textbf{Comparison on LOL-Blur dataset.} Quantitative results in Tab.~\ref{tab:com-lolblur} that we selected recent state-of-the-art methods, including DRBN~\cite{yang2020fidelity}, DeblurGAN-v2~\cite{kupyn2019deblurgan}, MIMO~\cite{cho2021rethinking}, NAFNet~\cite{chu2022nafssr}, LEDNet~\cite{zhou2022lednet}, Retinexforme, Restormer~\cite{zamir2022restormer}, VQCNIR~\cite{zou2024vqcnir}, DarkIR~\cite{feijoo2025darkir} and CIDNet. Our method achieves competitive performance while maintaining lightweight and efficient characteristics.

\noindent \textbf{Visual Comparisons.} Fig.~\ref{fig:visual_comparison} demonstrate our method's effectiveness against state-of-the-art techniques across four challenging extreme dark datasets, highlighting its superior enhancement quality in recovering fine details and preserving natural color balance in extremely dark conditions.

\begin{figure}[t]  
    \centering  
    \includegraphics[width=0.9\linewidth]{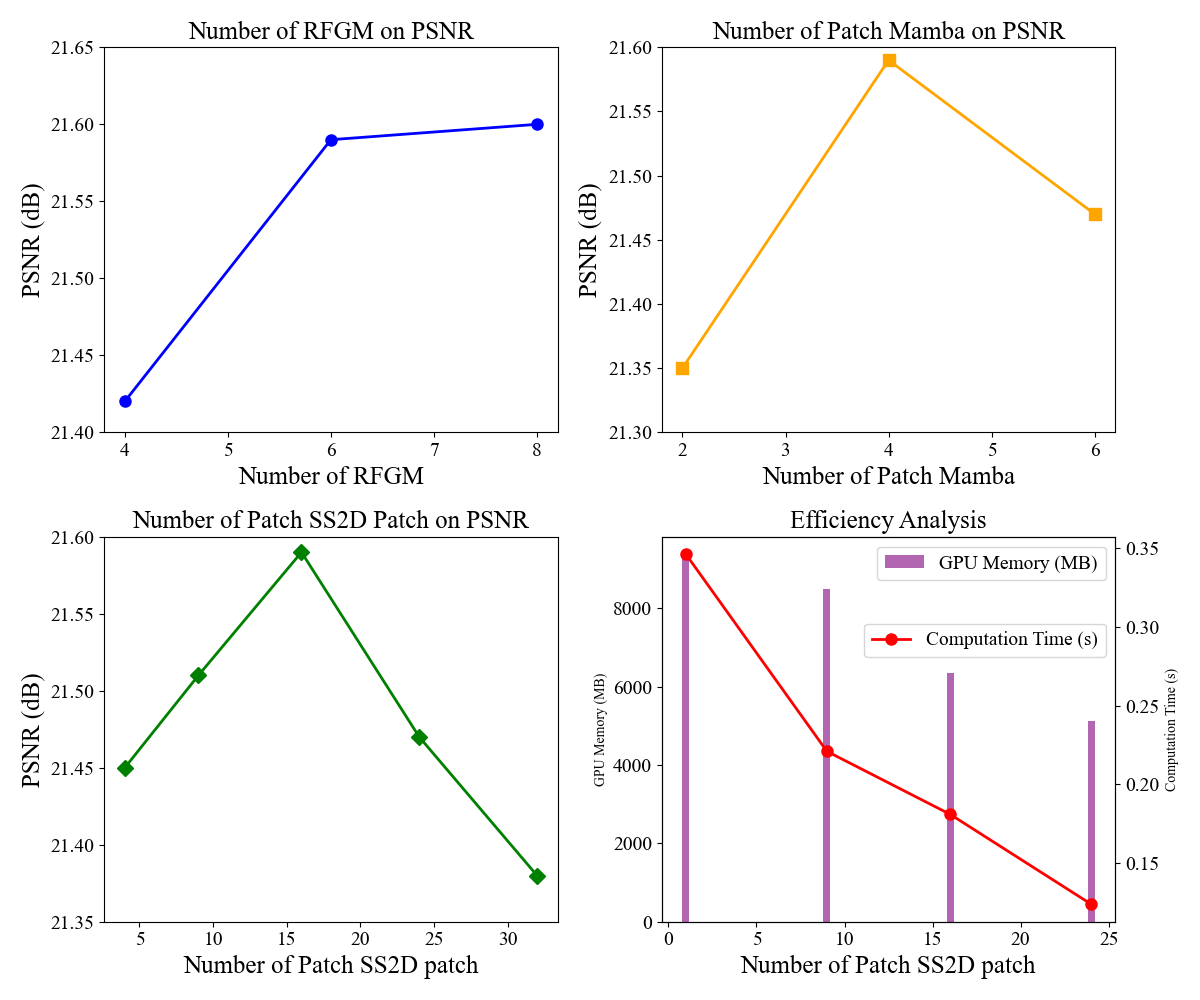}  
    \vspace{-0.2cm}
    \caption{Analysis of component counts and the efficiency of Patch SS2D. The plots illustrate the effects of varying the number of RFGM, Patch Mamba, and Patch SS2D patch on PSNR. Additionally, the efficiency analysis highlights the relationship between the number of patches, GPU memory usage, and computation time.}  
    \label{fig:ablation_study_num}  
    \vspace{-0.2cm}
\end{figure}

\begin{figure}[t]
    \centering
    \includegraphics[width=1.0\linewidth]{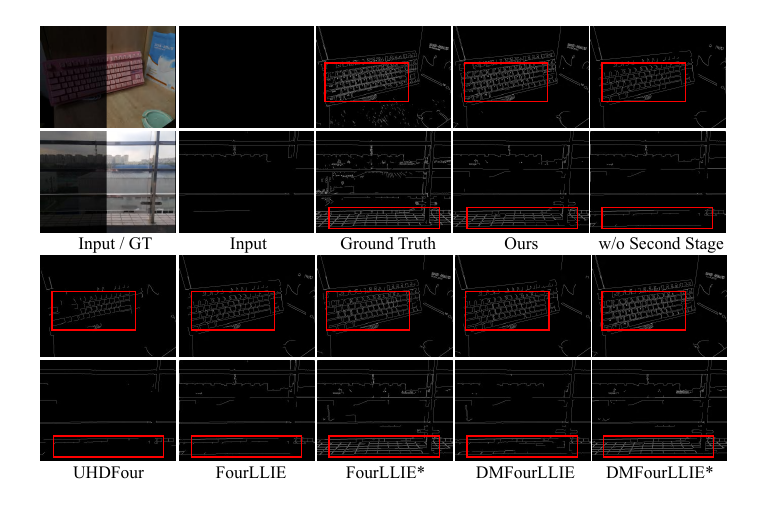}
    \vspace{-0.5cm}
    \caption{Canny edge detection visualization. The first column contains input and ground truth RGB images, followed by Canny edge results. We also assessed results without the second stage, confirming its role in enhancing edges and details. Special attention is given to the details in the red-boxed area; zooming in provides better visual contrast. In contrast to frequency-domain methods, which often struggle with fine detail preservation, we replaced the second stage of both FourLLIE* and DMFourLLIE* with our dual-branch Mamba, validating its efficacy in enhancing edge clarity. This comparison emphasizes the advantages of our approach in effectively retaining essential structural information.}
    \vspace{-0.2cm}
    \label{fig:canny}
\end{figure}

\begin{table}[t]  
\centering  
\renewcommand{\arraystretch}{1.2} 
\setlength{\tabcolsep}{4pt} 
\resizebox{1.0\columnwidth}{!}{ 
\begin{tabular}{l|ccccccc}  
\toprule  
Method & UHDFour & FourLLIE & Wave-Mamba &  MambaLLIE & Ours \\ \midrule 
H-Mean (CRAFT) & 0.4103 & 0.4211 & 0.4390 & 0.3684  & \textbf{0.4500} \\  
H-Mean (PAN) & 0.1760 & 0.1290 & 0.1760 & 0.1880  & \textbf{0.2350} \\  
\bottomrule  
\end{tabular}}  
\vspace{-0.2cm}
\caption{H-Mean comparison of text detection methods.}  
\label{tab:text_detection_H_mean}  
\end{table}

\subsection{Ablation Study}
\noindent \textbf{Component Ablation.}  To evaluate the contribution of each component in our proposed method, we conduct an ablation study by sequentially removing essential modules. The results are summarized in Tab.~\ref{tab:ablationbranch}. The removal of the second stage also leads to a noticeable decrease in metrics, indicating the importance of spatial information reconstruction. Each component has a measurable impact on performance, demonstrating the robustness of our approach.

\noindent \textbf{Component Count and Patch SS2D Efficiency Validation.}
Fig.~\ref{fig:ablation_study_num} presents an analysis of component counts and the efficiency of the Patch SS2D module. The plots demonstrate the effects of varying the number of RFGMs, Patch Mamba blocks, and Patch SS2D patches on PSNR. Specifically, the leftmost plot indicates that six RFGMs yield optimal PSNR results, while the second plot shows that four Patch Mamba blocks effectively balance performance. The third plot reveals that PSNR is maximized with 16 SS2D patches. The efficiency analysis in the bottom right plot illustrates the relationship between the number of patches, GPU memory usage, and computation time. As the number of patches increases from 1 to 24, both GPU memory consumption and computation time decrease significantly. Notably, using 16 patches strikes an optimal balance, reducing computation time by 47.7\% (from 0.346s to 0.181s) and GPU memory usage by 33.2\% (from 9362MB to 6252MB).

\noindent \textbf{Plug-and-Play Effectiveness of Core Modules.}
RFGM and Patch SS2D can be seamlessly integrated into existing Fourier-based methods with minimal additional parameters. Tab.~\ref{tab:plug_test} shows that incorporating our modules into FourLLIE and DMFourLLIE significantly enhances performance while maintaining high efficiency, validating their effectiveness.

\subsection{Downstream Application}
\noindent \textbf{Edge Detection.}
To validate our approach in edge detection, we refer to the results in Fig.~\ref{fig:canny}. The enhanced Canny edge outputs show significant improvement in preserving fine details and sharp boundaries, particularly in the red-boxed areas highlighting intricate structures. Comparing these results with the original images demonstrates that our dual-stage method effectively enhances edge definition, recovering details lost in extremely dark conditions.

\noindent \textbf{Text Detection.}
As shown in Tab.\ref{tab:text_detection_H_mean}, we conducted text detection on the LOL-Text dataset using the Text-CP~\cite{lin2025text}, alongside CRAFT~\cite{baek2019character} and PAN~\cite{wang2019efficient} detectors. The H-Mean, representing the mean of precision and recall, indicates a substantial improvement in our method's performance. This highlights our ability to recover details effectively in dark image conditions.

\section{Conclusion}
In this work, we presented an efficient dual-stage approach for recovering fine-grained details in extremely dark images, significantly improving edge and text detection performance. Our method utilizes a Residual Fourier-Guided Module and complementary Mamba modules, demonstrating robust enhancements in detail preservation.
Looking ahead, we will explore further avenues to ensure not only the recovery of intricate details but also the restoration of color semantic consistency in dark images, enhancing overall image quality and usability in practical applications.

\bibliography{aaai2026}



\end{document}